\documentclass[sigconf]{acmart}
\usepackage{epsfig,setspace,amsmath,epsf,bm,graphicx,float,mathtools}
\usepackage{algpseudocode}
\usepackage{dsfont}
\usepackage{subcaption}

\newtheorem{theorem}{\bf{Theorem}}
\newtheorem{corollary}{\bf{Corollary}}
\newtheorem{definition}{\bf{Definition}}
\newtheorem{remark}{\bf{Remark}}

\newtheorem{assumption}{\bf{Assumption}}
\newenvironment{Proof}[1]{\medskip\par\noindent{\bf Proof:\,}\,#1}{{\mbox{\,$\blacksquare$}\par}}

\newcommand{\st}{{\text{s.t.}}}
\newcommand{\e}{{\mathbb{E}}}
\newcommand{\ind}{{\mathds{1}}}
\allowdisplaybreaks

\AtBeginDocument{%
  \providecommand\BibTeX{{%
    \normalfont B\kern-0.5em{\scshape i\kern-0.25em b}\kern-0.8em\TeX}}}

\setcopyright{none}
\renewcommand\footnotetextcopyrightpermission[1]{} 
\acmDOI{}
\acmConference[]{}{}{}
\acmISBN{}
\begin{document}

\title{CAFe: Cost and Age aware Federated Learning}

\author{Sahan Liyanaarachchi}
    \affiliation{
    \institution{University of Maryland}
    \city{College Park}
    \country{USA}}
    \email{sahanl@umd.edu}

\author{Kanchana Thilakarathna}
    \affiliation{
    \institution{University of Sydney}
    \city{Sydney}
    \country{Australia}}
    \email{kanchana.thilakarathna@sydney.edu.au}

\author{Sennur Ulukus}
    \affiliation{
    \institution{University of Maryland}
    \city{College Park}
    \country{USA}}
    \email{ulukus@umd.edu}

\begin{abstract}
In  many federated learning (FL) models, a common strategy employed to ensure the progress in the training process, is to wait for at least $M$ clients out of the total $N$ clients to send back their local gradients based on a reporting deadline $T$, once the parameter server (PS) has broadcasted the global model. If enough clients do not report back within the deadline, the particular round is considered to be a failed round and the training round is restarted from scratch. If enough clients have responded back, the round is deemed successful and the local gradients of all the clients that responded back are used to update the global model. In either case, the clients that failed to report back an update within the deadline would have wasted their computational resources. Having a tighter deadline (small $T$) and waiting for a larger number of participating clients (large $M$) leads to a large number of failed rounds and therefore greater communication cost and computation resource wastage. However, having a larger $T$ leads to longer round durations whereas smaller $M$ may lead to noisy gradients. Therefore, there is a need to optimize the parameters $M$ and $T$ such that communication cost and the resource wastage is minimized while having an acceptable convergence rate. In this regard, {\bf \emph{we show that the average age of a client at the PS appears explicitly in the theoretical convergence bound,}} and therefore, can be used as a metric to quantify the convergence of the global model. We provide an analytical scheme to select the parameters $M$ and $T$ in this setting.
\end{abstract}


\maketitle

\pagestyle{empty}

\section{Introduction}
Federated learning (FL) was first introduced in \cite{McMahan} as a new machine learning paradigm where the task of model training is carried out in a decentralized structure so as to maintain data privacy among the participating entities. Since its emergence, it has been extensively used in a variety of real world applications such as banking applications, healthcare applications, human mobility predication and recommendation systems  where data privacy is critical \cite{pmf,fed_apps}. As such, FL is also increasingly becoming popular in mobile/IoT networks and mobile edge computing as a private and secure method of leveraging data produced by mobile/IoT devices.

A typical FL model consists of a parameter sever (PS) along with several clients each with a local dataset and their collective goal is to train the global model at the PS without explicitly revealing their local data to the PS. This is achieved through multiple communication/training rounds between the PS and the clients where at the beginning of each round, the PS broadcasts its model weights to the clients and then each client computes a local gradient based on the broadcasted global model. These local gradients are then sent back to the PS where they are aggregated and used to update the global model. This process is repeated until the global model at the PS has converged. This technique is often referred to as federated averaging (FedAvg) which was proposed in \cite{McMahan,fedavg} and shown to converge under the assumptions that the client participation is uniform and data is uniformly distributed among all the clients (i.i.d.~data distribution across clients). 

However, in many practical applications, the client datasets are non i.i.d., computation resources of clients are not uniform and network connectivity is unreliable, leading to non-uniform client participation. To this end, there have been many iterations of FedAvg in recent years. The convergence of the FedAvg algorithm under non i.i.d.~data distributions  was explicitly studied in \cite{fedavg_nidd} under the settings of full and partial client participation where the authors show that even under data heterogeneity FedAvg achieves convergence with an appropriately chosen decaying learning rate.

In addition to data heterogeneity among clients, another key challenge that arises in the FL models, is the presence of \emph{stragglers} among clients. If the FL model waits for all clients before updating the model, then the round duration would be dominated by the stragglers. One way to mitigate the straggler issue is to update the global model in an asynchronous manner \cite{async1,async2}. However, in this case, the aggregation of delayed gradients must be handled appropriately \cite{anarch}. Another simple but effective technique is to terminate the training rounds based on a reporting deadline \cite{dl1,dl2}.

As a result of many iterations and variants proposed over the years, it has become a significant challenge to configure a FL system to attain optimal performance with minimum resource wastage under a heterogeneous environment. It is common to select a portion of clients as learners for each round, and a reporting deadline to improve the efficiency of the system~\cite{refl}. System developers often rely on trial and error experimental strategies to determine appropriate values for these hyper-parameters. However, inaccurate hyper-parameters can lead to significant resource wastage and lower model performance~\cite{FedTune}. For example, if clients are not able to complete a local training round before the reporting deadline due to low computation resources of clients and/or network delays, scarce resources will be wasted at client devices. Especially in mobile/IoT networks, where every bit of resources at devices is paramount, such inefficiencies must be avoided. 
 
In this paper, we deduce fundamental relationships among model performance and resource heterogeneity with the aim of minimizing the resource wastage and communication cost, while maintaining an acceptable level of performance which often is determined by the convergence of model and the time for convergence. In order to quantify the performance of the FL framework, we propose to utilize the average age of clients (AoC) at PS. 

The average age of information (AoI) was introduced in \cite{rts2012} as a metric to measure the timeliness of communication systems \cite{age1,age2,age3}. The importance of timeliness in FL was addressed in \cite{buyukates_2021} where they consider exponential waiting times between client availability windows. This was later extended by \cite{mitra_2023} to an asynchronous hierarchical FL system. In this work, we show that through convergence analysis, for the FL model considered, AoC is in fact a direct quantifier for model performance. In particular, {\bf \emph{we show that the average AoC of users appears explicitly in the theoretical convergence bound,}} and therefore, can be used as a metric to quantify the convergence of the global model. That is, we show that lower average AoC leads to faster convergence. Therefore, the aim of our work is to find a good FL scheme that achieves timely communication with the clients and at the same time minimizes the resource wastage and communication costs incurred.

To summarize our contributions:
\begin{itemize}
    \item  
    We introduce age of clients (AoC) as a metric to quantify the model performance and show that when the total number clients is comparably larger than the number of learners in each training round, the average AoC at PS is directly related to the model performance in terms of the deviation from the optimal point (optimality gap) and time for convergence. In particular, we prove through theoretical convergence analysis and numerical validation, that lower AoC leads to better model performance.
    \item  
    We propose metrics to quantify resource wastage and communication costs and provide analytical expressions to compute them for a given FL model. These cost metrics along with the age metric are used to find optimal hyper-parameters $M$ and $T$ that are required to attain the desired levels of resource efficiency and model performance.
    \item
    We develop two resource efficient schemes: (i) age weighted update, and (ii) aggregated gradient update; that can be implemented in the presence of heterogeneous clients distributions. Through numerical results, we show that the introduced schemes help mitigate issues arising from biased or adversarial clients. Numerical results show that there is a significant performance gain when using age weighted updates compared to a conventional FedAvg update when the number of clients with biased datasets is high for the considered FL model.
\end{itemize}

\section{Related Work}
Recently, there has been growing interest in the domain of cost aware FL. \cite{cost_aware} considers the problem of jointly analyzing client association, transmission and computation optimization for a hierarchical FL (HFL) system with over the air computing where they decouple the joint problem into a client association subproblem and a resource optimization  subproblem and solve them via backward induction. In the REFL framework introduced in \cite{refl}, the goal is to jointly optimize resource efficiency along with the time to accuracy. They consider a reporting deadline based FL model with varying client availability periods and propose an intelligent participant selection algorithm to maximize the resource diversity of the FL training process. They also propose a staleness aware aggregation technique to utilize the updates of stragglers.

SAFA \cite{safa} and FLeet \cite{fleet}  are other schemes which implement a reporting deadline and aim to maximize the resource efficiency with the use of stale updates. SAFA collects all the updates it receives regardless of the round deadline and only applies updates that do not exceed a bounded staleness threshold. FLeet implements a staleness based dampening factor to utilize all the received updates. Oort \cite{oort} is another framework which studies the trade-off between faster training and resource diversity  where they employ  a participant selection algorithm to select faster learners. This minimizes the round duration at the expense of under-utilizing the datasets of slower clients.

Another problem of interest in this domain is the problem of  optimal hyper-parameter selection. FedTune \cite{FedTune} studies this problem and introduces an algorithm to select the optimal system parameters to optimize to computation time, transmission time, computation load and transmission load of the involved FL system. 

Age aware FL \cite{age-aware} has also been studied in the past, where the focus has been to design scheduling policies using age based metrics. \cite{arafa2020} introduces a new metric called age of update (AoU), which is similar to AoC in our work, and  uses it as a metric to design an age based scheduling policy which can increase the training efficiency. The importance of age in FL has been highlighted in all these prior works. However, age's direct link to the model performance has not been explored theoretically in them. Our work is the first to explore the fundamental link between the age and convergence in FL, together with cost aspects of FL. In fact, {\bf \emph{we show that age of a typical client is what controls the convergence rate of FL.}}

\begin{figure*}[ht!]
    \centering
    \includegraphics[width=2\columnwidth]{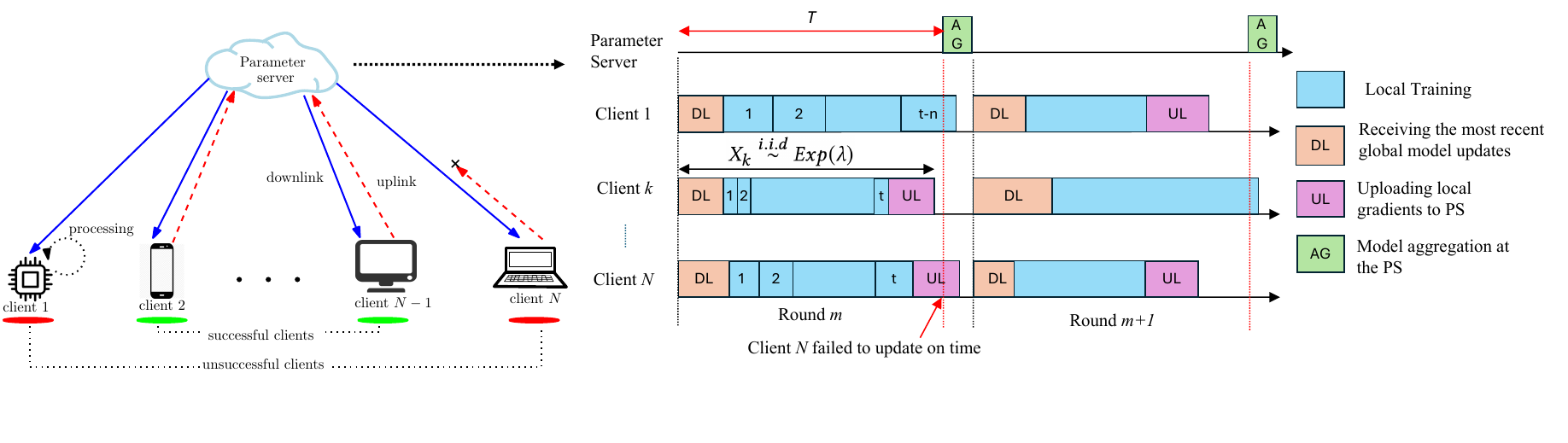}\vspace{-8mm}
    \caption{Federated learning (FL) model.}
    \label{fig:sys_model}
\end{figure*}

\section{System Model}\label{sec:sys}
Consider the FL setting shown in Figure \ref{fig:sys_model}, where the global model at the PS is trained using $N$ clients. At the beginning of  each training round $t$, the PS broadcasts its current global model weights $w_t$ to all the $N$ clients. Each client then uses these global model weights to train on their local data to produce localized gradients, which are then transmitted back to the PS for aggregation. We assume that for each client, the total time taken for down-link, up-link and the localized training is encapsulated by an exponential distribution with rate $\lambda$. Let $X_k$ represent the time from the beginning of the training round to the time at which the PS received the update from the  $k$th client. Then, $X_k \sim Exp(\lambda)$ and i.i.d.~over $k$.

In each training round, we assume the PS waits a time $T$, until at least $M$ clients have provided their local gradients before updating the global model. If the number of clients that sent back their local gradients by time $T$ is less than $M$, we say that the training round has failed. If a training round fails, the PS will discard all the received updates, re-broadcast the global model weights and restart the training round. If at least $M$ clients send back their local gradients by time $T$, then the PS will use the gradients of all the clients that responded back by time $T$ to update the global model.

Let the local dataset at client $k$ be represented by $D_k$. The goal of the PS is to minimize a loss function $l(\cdot)$ using the data distributed among the $N$ clients. This can be expressed as a distributed optimization problem as follows,
\begin{align}
    \min_w \left\{F(w) = \frac{1}{N}\sum_{k=1}^NF_k(w)\right\},
\end{align}
where $F_k(w) = \frac{N|D_k|}{\sum_i |D_i|}\sum_{d\in D_k}\frac{l(w,d)}{|D_k|}$ is the scaled local objective function associated with client $k$. 

Once the global model weights are received by client $k$, it tries compute the local gradient based on the function $F_k(w)$. This local gradient is often computed using a random mini-batch of the local dataset and hence is associated with a noise corresponding to the random sampling. Let the local gradient be represented by $\nabla \tilde F_k(w)$. These local stochastic gradients (SGD gradients) are then sent back to the PS where they are aggregated. Let $S_t$ be the set of clients that successfully sent back their updates (local gradients) in round $t$ before the deadline $T$. Then, the global model at the PS will be updated as follows,
\begin{align}\label{eqn:global_up}
     w_{t+1}= w_t - \frac{\eta_t \ind_{\{|S_t|\geq M\}} }{|S_t|}\sum_{k\in S_t}\nabla\tilde F_k(w_t),
\end{align}
where $\eta_t$ is the  associated learning rate in round $t$ and $\ind_{\{\cdot\}}$ is the indicator function. We term the FL scheme described above as the $M$-client update (MCU) scheme.

\section{Cost and Age Metrics}
In this section, we describe the metrics of interest for the FL model described in Section~\ref{sec:sys}.

\subsection{Resource Wastage}
We define the resource wastage $C_w$ as the total time spent on processing updates that were eventually discarded at the PS. Let $n$ represent the number of clients that responded back by $T$ in a particular training round. Then, $n \sim Binorm(N,p)$ where $p = 1-\exp(-\lambda T)$. In each training round, if $n<M$ we assume the round has failed and therefore all the received updates would be discarded. If $n\geq M$, then the PS would use all updates it received  during that particular training round, resulting in a wastage of resources  only  in the clients that failed to send their local gradients by time $T$. Under this scheme, the resource wastage in each round is as follows,
\begin{align}
    C_w =
    \begin{cases}
     NT, &  n < M,\\
        (N-n)T,  &  n \geq M.
    \end{cases}
\end{align}
Then, the average computational cost wasted $\e[C_w]$ is given by Theorem~\ref{thrm:cost}.

\begin{theorem}\label{thrm:cost}
Under the given federated learning (FL) model, the average resource wastage is given by,
    \begin{align}
        \e[C_w] = \frac{(1-p)NT+T\sum_{n=0}^{M-1}np_n}{(1-q)},
    \end{align}
where $p_n = {N \choose n}p^n(1-p)^{N-n}$ and $q=\sum_{n=0}^{M-1}p_n$. 
\end{theorem}

\begin{Proof}
 Let $K$ denote the number of training rounds until a successful round. Then, $K \sim Geom(q)$, and we have,
 \begin{align}
     \e[C_w]&=\e\left[(K-1)NT\right]+\e\left[(N-n)T|n\geq M\right]\nonumber\\
          &=\frac{q}{1-q}NT+\frac{\sum_{n=M}^N(N-n)Tp_n}{(1-q)} \nonumber\\
          &=\frac{q}{1-q}NT+NT-\frac{\sum_{n=M}^NnTp_n}{1-q}\nonumber\\
          &=\frac{NT}{1-q}+\frac{\sum_{n=0}^{M-1}nTp_n}{1-q}-\frac{\sum_{n=0}^NnTp_n}{1-q}\nonumber\\
          &= \frac{NT}{1-q}+\frac{\sum_{n=0}^{M-1}nTp_n}{1-q}-\frac{NTp}{1-q}\nonumber\\
          &= \frac{(1-p)NT+T\sum_{n=0}^{M-1}np_n}{(1-q)},
 \end{align}
 giving the desired expression.
\end{Proof}

\begin{corollary}\label{coro:cw}
For a fixed $T$, the optimal $M$ that minimizes $\e[C_w]$ is $M=1$.
\end{corollary}

\begin{Proof}
    For a fixed $T$, $p$ will be fixed. Thus, to minimize $\e[C_w]$, we need to minimize the partial expectation $\sum_{n=0}^{M-1}np_n$ and maximize $1-q$. Both of these are achieved simultaneously by $M=1$.
\end{Proof}

\subsection{Communication Cost}
Another metric of importance is the  communication cost incurred per successful update of the global model. This is denoted by $C_b$ and is defined to be equal to the number of communication rounds required for a successful training round. Then, the average communication cost $\e[C_b]$ isgiven by,
\begin{align}\label{eqn:com_cost}
    \e[C_b] = \e[K]= \frac{1}{1-q},
\end{align}
where $K$ and $q$ are as defined in the proof of Theorem~\ref{thrm:cost}. Following similar arguments as in the proof of Theorem~\ref{thrm:cost}, for a fixed $T$, choosing $M=1$ minimizes the average communication cost.

\subsection{Average of Age of the Client}
\begin{figure}
     \centering
     \includegraphics[width=\columnwidth]{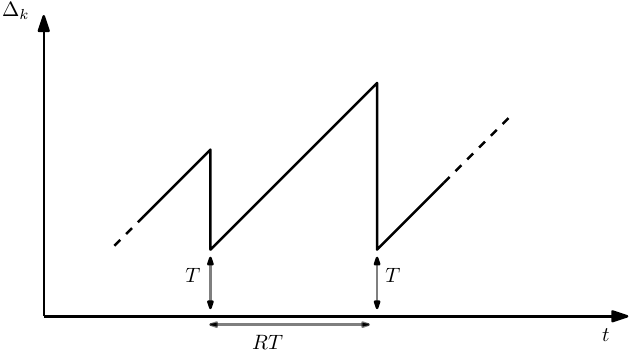}
     \caption{The evolution of the instantaneous age of the $k$th client with time.}
     \label{fig:age_cycle}
\end{figure}

The average age of the client is the time average of the associated AoI process corresponding to the client updates. In particular, the age of a client $k$ denoted by $\Delta_k$, increases linearly until the client under consideration participates in a successful round and provides its local gradient to the PS. Once a successful update from the client is received, then $\Delta_k$ will drop to $T$. This is based on the assumption that we update the global model only at the end of the training round while the local training is assumed to start at the beginning of a training round. In particular, for every failed  training round, the round duration would be $T$ and for a successful training round, even if all $N$ clients managed to send updates before time $T$, we would update the global model only after $T$ has elapsed (for $N\gg 2M$ probability that all clients finish before $T$ is very small and therefore round duration would be $T$ almost surely). Hence, the age drops to $T$ once an update has been successfully received. Figure \ref{fig:age_cycle} shows how the age of the $k$th client evolves under the above assumptions.

Since we consider $N$ homogeneous clients (i.e., the same update rate $\lambda$), we only need to consider the age of one particular client. Under the above assumptions, Theorem \ref{thrm:aoi} gives the average age of a client at the PS.

\begin{theorem} \label{thrm:aoi}
    For a finite $T$, the time averaged age of the client at the PS is given by,
    \begin{align}
        \e[\Delta_k] = \frac{T}{2}+\frac{T}{\sum_{n=M-1}^{N-1}{N-1 \choose n}p^{n+1}(1-p)^{N-1-n}}.
    \end{align}
\end{theorem}

\begin{Proof}
    Let $\tilde{n}$ represent the number of successful updates from the other $N-1$ clients. Then, $\tilde{n}\sim Binorm(N-1,p)$. The age of the client $k$ would decrease only if it participated in a successful training round. The probability that the client $k$ participates in a round is $p$ and the probability that the round is successful is equal to the probability that at least $M-1$ clients from the remaining $N-1$ clients participated in the same round. Therefore, the probability that client $k$ participates in a successful round is $pPr(\tilde{n}\geq M-1)$. Let $R$, represent the number of training rounds till a successful update is received from client $k$. Then, $R\sim Geom(r)$ where $r= 1-pPr(\tilde{n}\geq M-1)$. Then, the average age of client $k$, denoted by $\e[\Delta_k]$, can be obtained by computing the ratio between the average area under the age graph and the mean time between two successive drops in the age graph, as follows,
    \begin{align}
        \e[\Delta_k]&=\frac{\e[(2T+RT)RT]}{2\e[R]T}\nonumber\\
                    & = \frac{2\e[R]T^2+\e[R^2]T^2}{2\e[R]T}\nonumber\\
                    & = T+\frac{\e[R^2]T}{2\e[R]}\nonumber\\
                    & = T+\frac{(1+r)T}{2(1-r)} \nonumber\\
                    &= \frac{T}{2}+\frac{T}{1-r}. \label{eqn:age_r}
    \end{align}
Substituting $Pr(\tilde{n}\geq M-1) = \sum_{n=M-1}^{N-1}{N-1 \choose n}p^{n}(1-p)^{N-1-n}$ in \eqref{eqn:age_r}  yields the desired result.
\end{Proof}

\begin{corollary}\label{coro:age}
    For a fixed $T$, the optimal $M$ that minimizes the average age of the client is $M=1$.
\end{corollary}

\begin{Proof}
    To minimize $\e[\Delta_k]$, we need to maximize the term $pPr(\tilde{n}\geq M-1)$, which is achieved by setting $M=1$.
\end{Proof}

\begin{remark}
    Though at a first thought, the average age of the client may seem a relatively unrelated metric for FL, in the next section, we will show through a rigorous convergence analysis that $\e[\Delta_k]$ is directly associated with the rate of convergence of the model for the considered FL scheme.
\end{remark}

\begin{remark}
    When we set $M=1$, this corresponds to the situation where the PS uses the local gradients of all the clients it has received by time $T$, to update the  global model. In this case, if only a few clients responded, then the net gradient vector used to update the global model would be noisy.
\end{remark}

\section{Convergence Analysis}
In this section, we provide the convergence analysis of the MCU scheme described in Section~\ref{sec:sys}. We closely follow the outline of the convergence analysis in \cite{fedavg_nidd} and use the following assumptions to proceed in our analysis.
 
\begin{assumption}\label{assum:smooth}
    The loss functions at all clients are $L$-smooth. For any $x,y$, the loss function at client $k$ satisfies,
    \begin{align} 
        F_k(y)\leq F_k(x)+(y-x)^T\nabla F_k(x) + \frac{L}{2}\|y-x\|^2.
    \end{align}
\end{assumption}

\begin{assumption}\label{assum:conv}
    The loss functions at all clients are $\mu$-strongly convex. For any $x,y$, the loss function at client $k$ satisfies,
    \begin{align} 
        F_k(y)\geq F_k(x)+(y-x)^T\nabla F_k(x) + \frac{\mu}{2}\|y-x\|^2.
    \end{align}
\end{assumption}

\begin{assumption}\label{assum:ube}
    The SGD gradient at each client is an unbiased estimator of its actual gradient with respect to the client dataset and has bounded variance,
    \begin{align}
    \e[\nabla\tilde{F}_k(x)|x]&=\nabla F_k(x),\\
    \e[\|\nabla F_k(x)-\nabla\tilde{F}_k(x)\|]&\leq \sigma^2,
    \end{align}
    where $\nabla F_k(x)$ is the actual gradient at the $k$th client. 
\end{assumption}
\begin{assumption}\label{assum:ind}
    The SGD gradients between clients are independent. For all $k,l \in \{1, 2, \dots , N\}$, and model $x$,
    $(\nabla F_k(x)- \nabla \tilde F_k(x)) \perp (\nabla F_l(x)- \nabla \tilde F_l(x))$.
\end{assumption}

\begin{definition}
    The heterogeneity of the data distribution across the clients is given by $\Gamma = F^*-\frac{1}{N}\sum_{k=1}^N F_k^*$, where $F^*$ is the global optimum and $F^*_k$ is the local optimum with respect to  $D_k$.
\end{definition}

\begin{remark}\label{rem:data}
    Since the terms $ \frac{N|D_k|}{\sum_i |D_i|}$ have been absorbed into the loss function $F_k(w)$, $\sigma^2$ in Assumption~\ref{assum:ube} is proportional to $max_k \frac{N|D_k|}{\sum_i |D_i|}$. Therefore, if we have have a highly non uniform distribution of samples among clients, $\sigma^2$ would also increase.
\end{remark}

Under the above assumptions, Theorem~\ref{thrm:mcu_conv} below gives a bound on the expected deviation from the global optimum after $t$ training rounds for the MCU scheme.

\begin{theorem}\label{thrm:mcu_conv}
    Under the MCU scheme, the expected optimality gap after $t$ iterations satisfies the following bound,
    \begin{align}\label{eqn:mcu_conv}
        \e[F(w_t)]-F^*
        \leq \frac{L}{(2L+\tilde S N\mu t)}\left(\frac{(4L\Gamma+\frac{2\sigma^2}{M})}{\mu}+(L+N\mu\tilde S)\varepsilon_1\right),
    \end{align}
    where $\tilde{S}= \sum_{n=M-1}^{N-1}\frac{1}{n+1}{N-1 \choose n}p^{n+1}(1-p)^{N-1-n}$, $\varepsilon_1= \e\|w_1-w^*\|^2$, and  $w^*$ is the global optimal model.
\end{theorem}

The proof of Theorem~\ref{thrm:mcu_conv} is given in Appendix~\ref{app:conv}. We have the following corollary.

\begin{corollary}\label{col:age_b}
    Under the MCU scheme, the convergence bound is $O\left(\frac{1}{tpPr(\tilde n \geq M-1)}\right)$, where $\tilde n \sim Binorm(N-1,p)$. Alternatively, this bound is equivalent to $O\left(\frac{\e[\Delta_k]}{Tt}\right)$.
\end{corollary}

\begin{Proof}
    For large $t$ the inequality \eqref{eqn:mcu_conv} reduces to the following,
    \begin{align}\label{col:age_1}
        \e[F(w_t)]-F^*
        &\leq \frac{L}{\tilde S N\mu t}\left(\frac{(4L\Gamma+\frac{2\sigma^2}{M})}{\mu}+(L+N\mu\tilde S)\varepsilon_1\right)\nonumber\\
        &\leq \frac{L\left(4L\Gamma+2\frac{\sigma^2}{M}+L\mu\varepsilon_1
        \right)}{\tilde S N\mu^2 t}+\frac{L\varepsilon_1}{t}.
    \end{align}
    Moreover, $\tilde S $ can be bounded as follows,
    \begin{align}\label{col:age_2}
        \tilde S &= \e\left[ \frac{\ind_{\{k \in S_t,|S_t|\geq M\}}}{|S_t|}\right]\nonumber\\
                &\geq \frac{1}{N}\e\left[\ind_{\{k \in S_t,|S_t|\geq M\}}\right]\nonumber\\
                &=\frac{pPr(\tilde n \geq M-1)}{N}.
    \end{align}
Combining \eqref{col:age_1} and \eqref{col:age_2}, we obtain,
\begin{align}
     \e[F(w_t)]-F^*\leq \max\left\{\frac{L}{\mu},1\right\}\frac{\left(4L\Gamma+2\frac{\sigma^2}{M}+L\mu\varepsilon_1
        \right)}{p Pr(\tilde n \geq M-1) \mu t}+\frac{L \mu \varepsilon_1}{\mu t}. \label{eqn:bound_2}
\end{align}
Since $pPr(\tilde n \geq M-1)\leq 1$, the first term dominates the upper bound proving the desired convergence bound. The age related bound is obtained by noticing 
 that the term $1-r =pPr(\tilde n \geq M-1) $ in \eqref{eqn:age_r}.
\end{Proof}

\section{Optimal design parameters}\label{sec:opt_params}
For the MCU scheme, we consider, the least number of learners required $M$, and the  reporting deadline $T$ as controllable design parameters of our model. According  to Corollary~\ref{coro:cw} and \eqref{eqn:com_cost}, we notice that setting $M=1$ minimizes both the resource wastage and the communication cost. When the data distribution across clients is non i.i.d.~and $\sigma^2$ is small, Corollary~\ref{col:age_b} shows that the optimality gap is bounded by $\frac{\e[\Delta_k]}{T}$. Therefore, we can use $\frac{\e[\Delta_k]}{T}$ as a metric to quantify the accuracy of the model. This would imply that having a large $T$ would be beneficial for the model to converge faster (i.e., the number of training rounds required is lower). However, this does not take into account the duration of the training rounds. Therefore, we choose to use $\e[\Delta_k]$ as a metric which encapsulates both the model accuracy and the round duration into account. According to Corollary~\ref{coro:age}, setting $M=1$ minimizes the age, and therefore leads to faster convergence. Thus, we choose $M=1$ as the optimal design parameter in this setting. Therefore, what remains is to find the optimal report deadline $T$, that minimizes the computational cost wastage, communication cost and age of clients.

\begin{remark}\label{rem:iid_conv}
    When the variance of the stochastic gradient $\sigma^2$ is high (compared to $4L\Gamma$ and $L\mu \varepsilon_1 $) , setting $M=1$ would not necessarily mean faster convergence. In particular, then the convergence rate is inversely proportional to the product $MPr(\tilde n \geq M-1).$ 
\end{remark}

\begin{figure}
     \centering
     \includegraphics[width = 0.9\columnwidth, trim={0 0 0 1cm}]{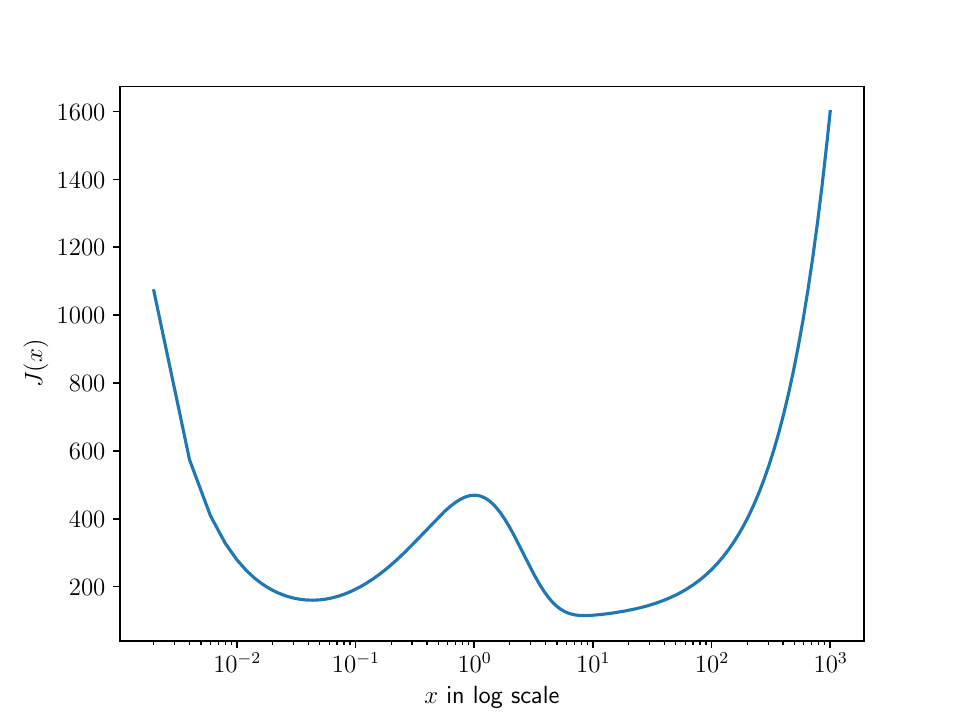}\vspace{-2mm}
     \caption{$J(x)$ for $N=50$, $\lambda =1$, $\alpha_w = 20$ and $\alpha_b=100$. \vspace{-2mm}}
     \label{fig:J_log}
 \end{figure}

Next, we scalarize our multi-objective optimization problem as follows. Let $\alpha_w$ and $\alpha_b$ denote the relative weights associated with $\e[C_w]$ and $\e[C_b]$ with respect to $\e[\Delta_k]$. Then, the goal is to minimize the following objective function $J(\lambda,T)$,
\begin{align}
    J(\lambda,T) = &\alpha_w \e[C_w]+\alpha_b \e[C_b] +\e[\Delta_k] \nonumber\\
    = & \alpha_w \frac{NTe^{-\lambda T}}{1-e^{-N\lambda T}} + \frac{\alpha_b}{1-e^{-N\lambda T}} + \frac{\lambda T}{\lambda}\left(\frac{1}{2}+\frac{1}{1-e^{-\lambda T}}\right).\label{eqn:mobj}
\end{align}
Taking $x = \lambda T$ in \eqref{eqn:mobj} yields the following optimization problem, 
\begin{align}
    \min_x & \quad J(x) \nonumber\\
    \st & \quad x \geq 0, \label{opt-prob}
\end{align}
where $J(x) = \alpha_w \frac{Nxe^{-x}}{\lambda(1-e^{-Nx})} + \frac{\alpha_b}{1-e^{-Nx}}+\frac{x}{\lambda} \left(\frac{1}{2}+\frac{1}{1-e^{-x}}\right)$. Unfortunately, \eqref{opt-prob} is not a convex optimization problem. However, the second and third term are convex functions and the first term is a bounded function for $x\geq0$. Therefore, for small values of $\alpha_w$, this can be approximated as a convex optimization problem. Moreover, as $x \to 0$ the second term goes to $\infty$ and as $x \to \infty$ the third term goes to $\infty$. Therefore, even for large values of $\alpha_w$, we can simply do an exhaustive search 
 within a reasonable bounded space to find the optimal point. Figure \ref{fig:J_log} depicts the typical variation of $J(x)$ with $x$.
 
\section{Heterogeneous Client Schemes}
In the MCU scheme, setting $M=1$ in general minimizes the resource wastage while having a faster convergence rate. However, this is under the assumption that the clients are homogeneous. If the clients are heterogeneous with respect to their processing power, setting $M=1$ would not be ideal. This is mainly due to the fact that the PS would be dominated by faster clients. This would be especially degrading in cases where the faster clients have bad datasets or in the event that the faster clients are adversarial in nature. Therefore, setting $M>1$ would be beneficial in these cases. However, if we adhere to the MCU scheme, this would come at the cost of increased resource wastage and communication costs. Therefore, we propose two schemes that would help mitigate the issues that would arise with heterogeneous clients.

\subsection{Age Weighted Update (AWU)} 
When a certain client has not updated the model for several rounds, the global model tends to drift away from the optimal with respect to the dataset of that particular client. In that case, giving more weight to client that has not updated the model in a long time would seem beneficial. Moreover, if a certain client always updates our model compared to others, the global model tends to get biased towards the local dataset of  that particular client, which can  also be detrimental to the system if the client dataset is not good or if the client is an adversary. Therefore, in this setting we update the model by weighting the client gradients based on a bounded increasing function of the age of the client at the PS. Moreover, we set $M=1$ so as to allow the global model to be updated in each round and increase the resource efficiency. In this scheme, the global model is updated as follows,
 \begin{align}
     w_{t+1}= w_t - \frac{\eta_t \ind_{\{|S_t|\geq 1\}} }{\sum_{j\in S_t}Q(\Delta_j)}\sum_{k\in S_t}Q(\Delta_k)\nabla\tilde F_k(w_t),
 \end{align}
where $\Delta_k$ is the age of the $k$th client at the PS and $Q(\cdot)$ is a positive bounded increasing function.

\subsection{Aggregated Gradient Update (AGU)} 
In the presence of heterogeneous clients, operating with $M>1$ in the MCU scheme can help the PS from being overwhelmed by faster clients. However, this would increase the number of failed rounds, and discarding the gradients that have been computed in the failed rounds would be a waste of resources. Therefore, in this scheme, we aggregate the computed gradients across failed rounds and update the global model based on the aggregated gradients received when a round is successful. In this scheme, the global model is updated as follows,
\begin{align}
     w_{t+1}= w_t - \frac{\eta_t \ind_{\{|S_t|\geq M\}} }{|S_t|}\sum_{k\in S_t}\nabla\bar F_k(w_t),
 \end{align}
where $\nabla\bar F_k(w_t)$ is the accumulated stochastic gradient of the $k$th client across the failed rounds until a successful round occurs. When a round is successful, the gradients are re-accumulated across failed rounds for the new global model. In other words, on the event of a failed round, the successful client would not restart the local training from the current global model but rather would continue its local training starting from its new local model. When a round is successful, all the local models will be re-synchronized with the global model.

\section{Numerical Results}
In this section, we validate our theoretical results through several experiments. 

\noindent\textbf{Experimental settings.} For simulations, we use MNIST dataset which consists of 10 classes, each corresponding to a different handwritten digit \cite{mnist}. Out of the total 60000 data points in the MNIST dataset, 50000 are used for the training set and 10000 is used for the test set. In each of the experiments, we consider 100 clients ($N=100$) where each client is allocated a local dataset by sampling the training set of the MNIST dataset appropriately. For each experiment, we use a simpler 4 layer perceptron model. To evaluate the model performance, in each of the experiments, we train the global model at least 1000 training rounds to achieve stable results and evaluate its accuracy on the test set of the MNIST dataset. The learning rate of each client is set at $\lambda =1$ in all experiments. To simulate the local training  time of clients, at the beginning of each training round, we obtain a random sample from $Exp(\lambda)$ distribution for each client. If  there are at least $M$  clients whose associated random sample is less than $T$, then we allow those clients to update the global model. Otherwise, the round is assumed to be failed and a new training round is started from the beginning.

\subsection{I.I.D.~vs Non I.I.D.~Data Distribution}
\begin{figure}
    \centering
    \includegraphics[width=\columnwidth]{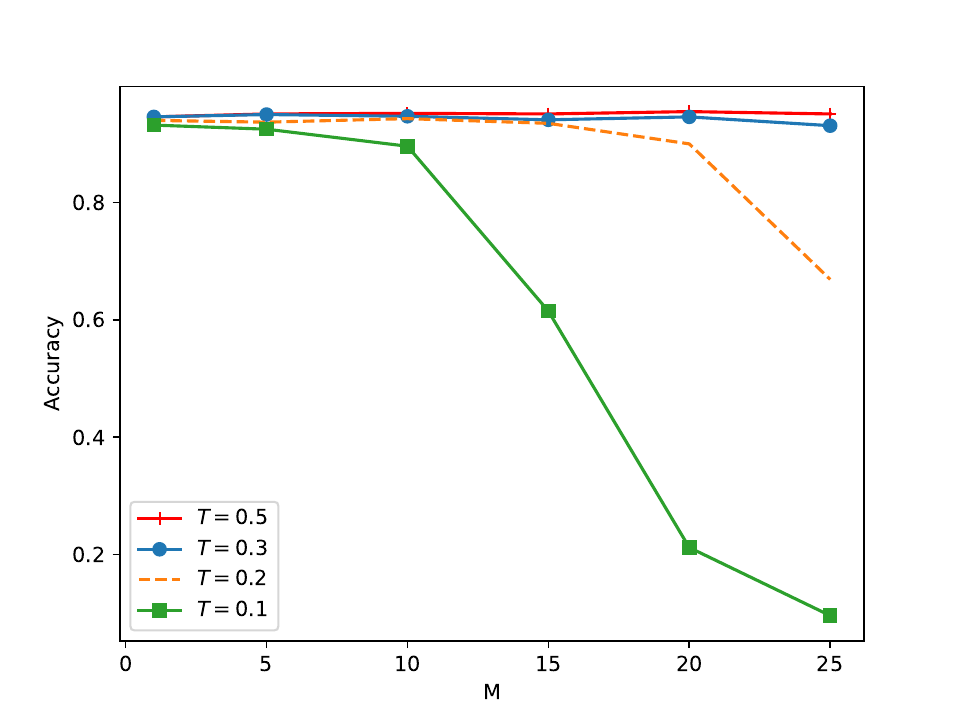}\vspace{-4mm}
    \caption{Variation of accuracy with $M$ for non i.i.d.~MNIST dataset after 1000 training rounds.}
    \label{fig:niid_vs_M}
\end{figure}

In the first numerical experiment, we compare how the accuracy varies with $M$ for a given value of $T$ when the data distribution among clients is non i.i.d.~in nature. For each client, we assign 3 classes by allocating a unique combination of classes from among $N \choose 3$ ways of selecting 3 classes out of 10. Further, each client is allocated an equal number of data points so as to achieve a small $\sigma^2$ value. Figure~\ref{fig:niid_vs_M} shows that by selecting $M=1$ in general yields a higher accuracy  compared to larger values of $M$. This increase in accuracy becomes more prominent as $T$ increases. This is to be expected since we have intentionally distributed data so as to increase the $\Gamma$ while maintaining a small value for $\sigma^2$. From \eqref{eqn:bound_2} it is evident that the optimality gap will be inversely proportional to $Pr(\tilde n \geq M-1)$. Hence, setting $M=1$ would lead to higher accuracy.

\begin{figure}
    \centering
    \includegraphics[width=\columnwidth]{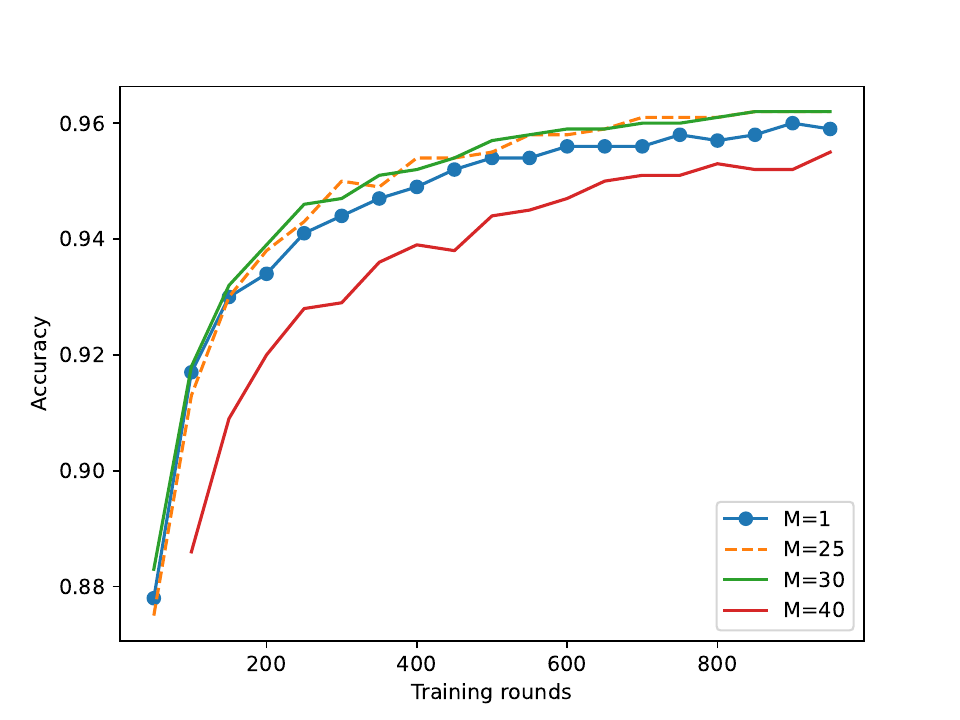}\vspace{-3mm}
    \caption{Variation of the accuracy with the number of training rounds for different values of $M$ for i.i.d.~MNIST dataset and $T=0.5$.}%\vspace{-4mm}
    \label{fig:iid_vs_M}
\end{figure}

\begin{figure}
    \centering
    \includegraphics[width=\columnwidth]{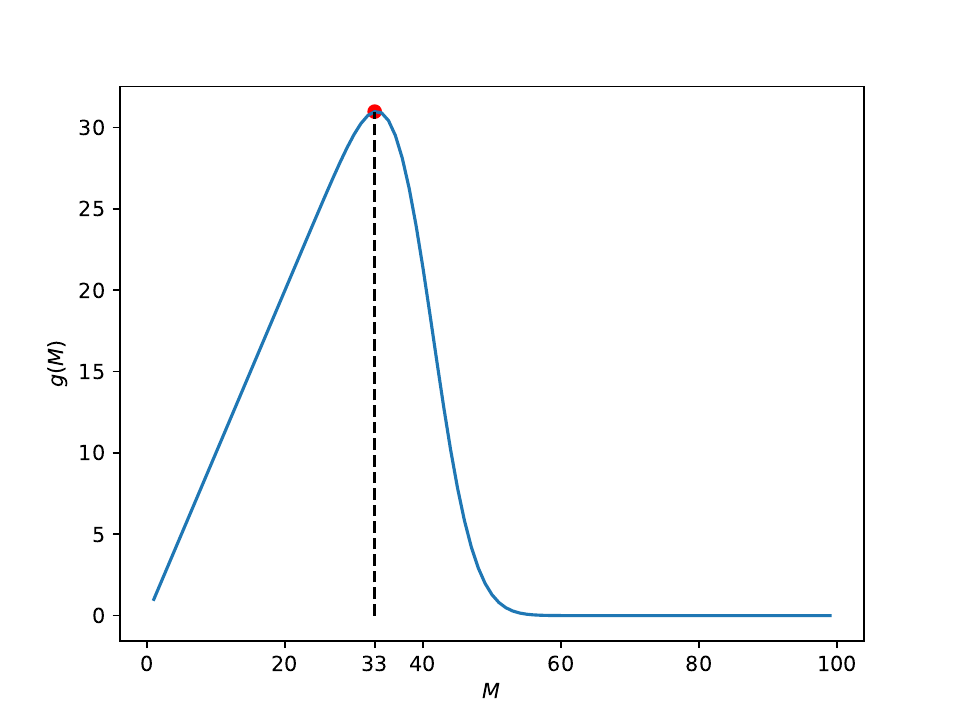}
    \caption{Variation of the $g(M)=MPr(\tilde n\geq M-1)$ with $M$ for $T=0.5$.}%\vspace{-4mm}
    \label{fig:iid_rate}
\end{figure}

Next, we repeat the same experiment as above, when the client datasets are sampled in an i.i.d.~manner. In this experiment, each client is allocated an equal number of randomly sampled data points from each of the 10 classes. We note that for some experiments in this case, setting $M=1$ would not be  optimal for comparatively larger values of $T$. Figure~\ref{fig:iid_vs_M} shows one such scenario. As seen in Figure~\ref{fig:iid_vs_M}, when $M$ increases from $M=1$ to $M=30$, the accuracy has increased but as $M$ increases further the accuracy has decreased. This behavior is similar in nature to the behavior predicted in Remark~\ref{rem:iid_conv}. If the initial deviation $\varepsilon_1$ and $\Gamma$ is comparatively small, then the model performance will be  proportional to $MPr(\tilde n\geq M-1)$. Figure~\ref{fig:iid_rate} shows $MPr(\tilde n\geq M-1)$ will achieve its maximum around $M=33$ and will decrease rapidly as it comes closer to $M=40$. Therefore, we believe that in this experiment the ideal conditions were met to achieve the predicted behavior. However, this behavior may not be observable for all i.i.d.~distributions as it depends on $\varepsilon_1$ as well.

\begin{figure}[t]
    \centering
    \includegraphics[width=\columnwidth]{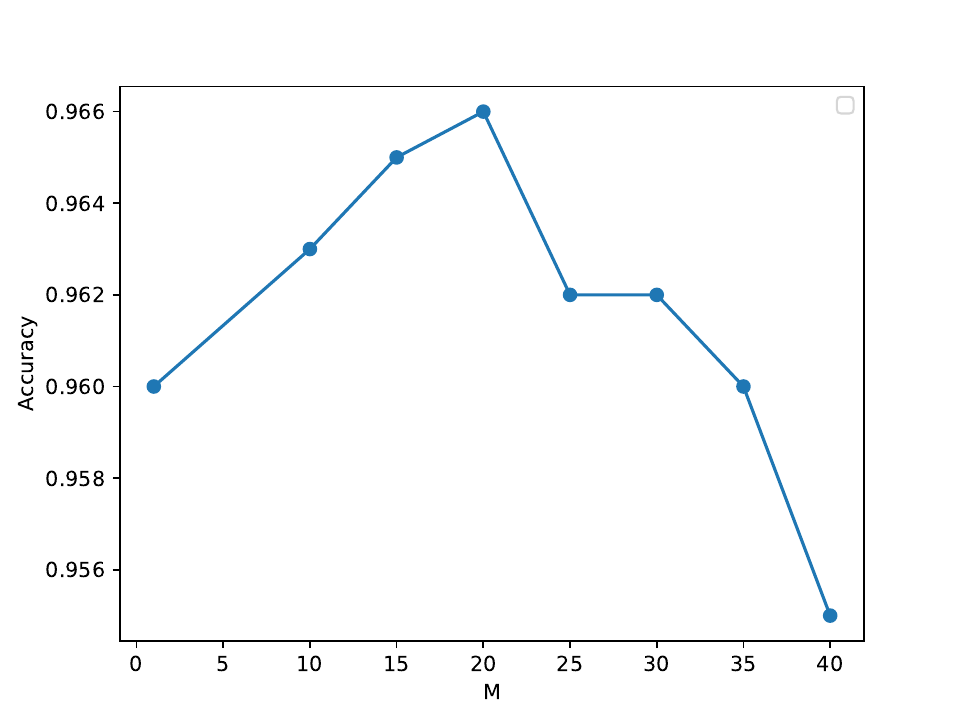}
    \caption{Variation of the accuracy with $M$ for i.i.d.~MNIST dataset after 1000 training rounds for $T=0.5$.}
    \label{fig:peak_iid_M}
\end{figure}

Moreover, even if we distribute the classes to the clients in a non i.i.d.~manner, if the number of data points allocated for each client is significantly different, then again we start to observe the above behavior for larger $T$ values. This is due to the fact that $\sigma^2$ is directly related to how the data points are distributed among the clients as suggested by Remark~\ref{rem:data}. In general, the graph of $MPr(\tilde n\geq M-1)$ can be used as a good indicator to determine whether $M>1$ would significantly degrade the model accuracy for a given $T$. However, it is not a guaranteed indicator of which $M$ value would yield the maximum accuracy due to various assumptions used in the proofs (see Figure~\ref{fig:peak_iid_M}). Additionally, even in cases where $M>1$ yields a higher accuracy, the improvement over the case when $M=1$ is very small. Therefore, $M=1$ is good choice for the minimum numbers of learners required per round.

\subsection{Resource Efficiency vs Accuracy}
As seen from the previous section, even though there may be instances where the model accuracy achieves its peak for a larger $M$, the difference is very insignificant compared to the instance $M=1$ (see Figure~\ref{fig:iid_vs_M}). Therefore, as suggested in Section~\ref{sec:opt_params}, we set $M=1$ and explore how the resource wastage, communication cost, normalized age ${\e[\Delta_k]}/{T}$ are linked to the parameter $T$. In this experiment, each client is allocated a single class (client $k$ is allocated class $k$ mod 10 ) and an arbitrary number data points are sampled. 

\begin{figure*}[t]
    \centering
     \begin{subfigure}[b]{0.9\columnwidth}
         \centering
            \includegraphics[width=\columnwidth]{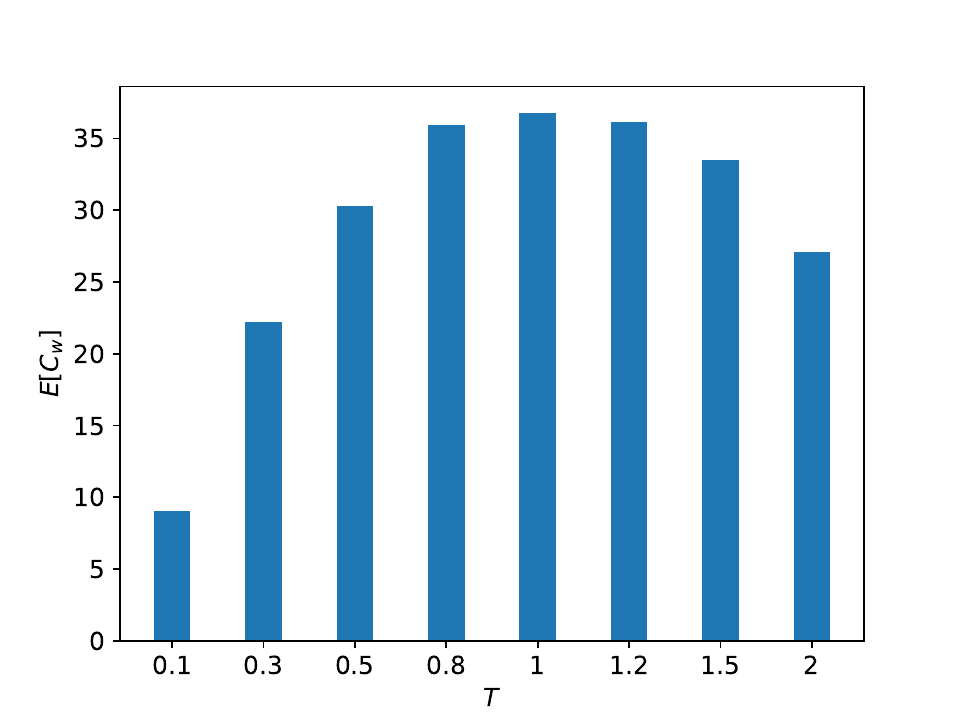}\vspace{-3mm}
         \caption{Resource wastage}\label{fig:rw}
     \end{subfigure}
     \hspace{5mm}
    \begin{subfigure}[b]{0.9\columnwidth}
         \centering
         \includegraphics[width=\columnwidth]{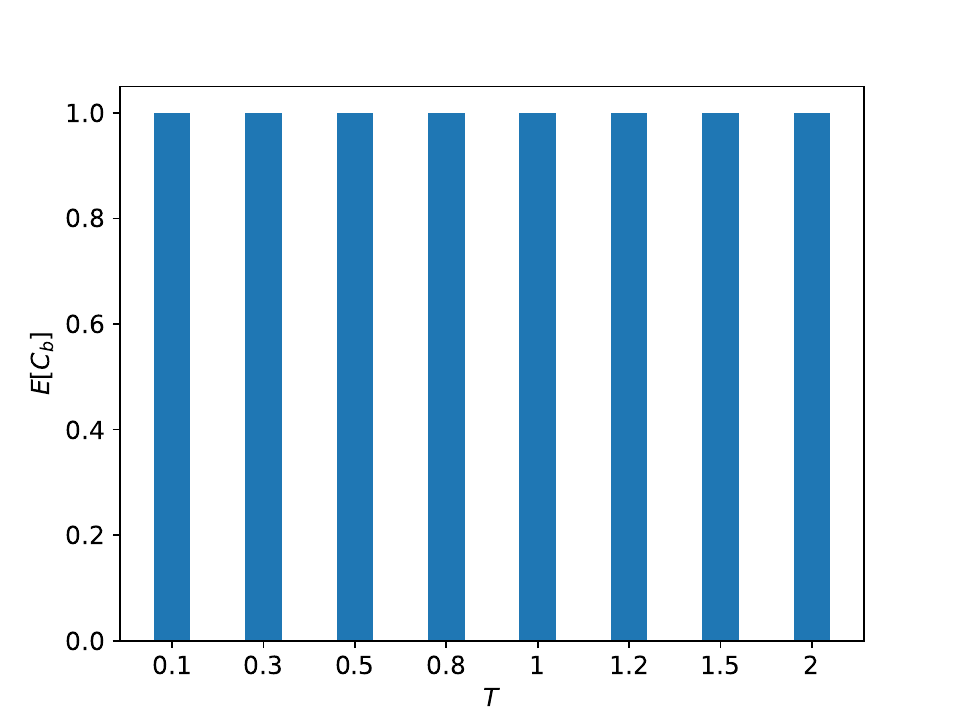}\vspace{-3mm}
         \caption{Communication cost}\label{fig:cc}
     \end{subfigure}
     \begin{subfigure}[b]{0.9\columnwidth}
         \centering
        \includegraphics[width=\columnwidth]{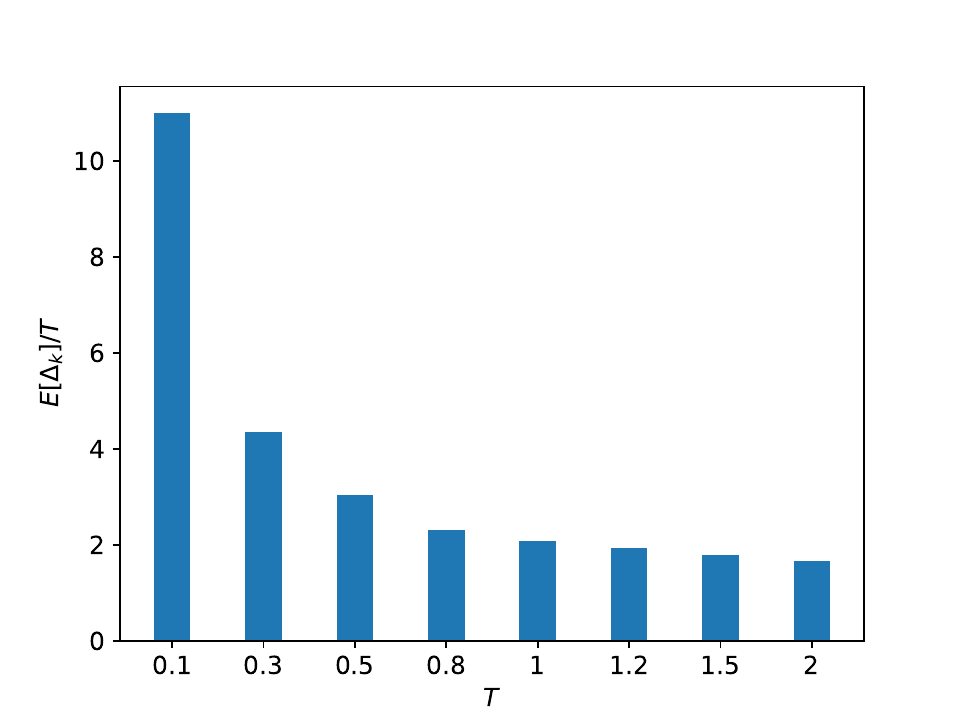}\vspace{-3mm}
         \caption{Normalized age}\label{fig:na}
     \end{subfigure}
      \hspace{5mm}
    \begin{subfigure}[b]{0.9\columnwidth}
         \centering
         \includegraphics[width=\columnwidth]{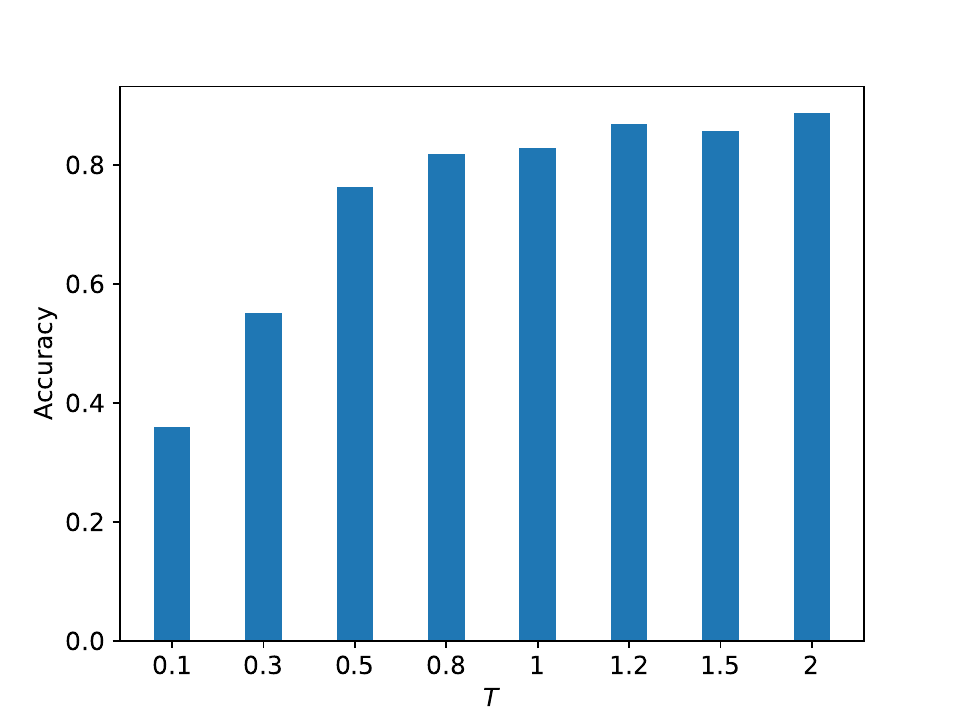}\vspace{-3mm}
         \caption{Accuracy} \label{fig:ac}
     \end{subfigure}\vspace{-3mm}
    \caption{Variation of resource wastage, communication cost, normalized age and accuracy with $T$ for $M=1$.}
    \label{fig:T_graphs}
\end{figure*}

Figure~\ref{fig:T_graphs} shows the variation of the quantities of interest with $T$. We see that the communication cost remains almost at 1 in each of the experiments as shown in Figure~\ref{fig:cc}. This is due to the fact that we are using a large number of clients. Therefore, almost every training round is a successful round for the considered $T$ values. We observe that the model accuracy increases with $T$ (see Figure~\ref{fig:ac}) while the normalized age decreases with $T$ (see Figure~\ref{fig:na}). This validates the result given by Corollary~\ref{col:age_b}. We also note that as $T$ increases, the resource wastage, first increases and then starts to decrease (see Figure~\ref{fig:rw}). This is because, as $T$ increases, more time is wasted by failed clients, resulting in an increase in resource wastage. But as $T$ increases further, the number of clients that fail also decreases, resulting in a decrease in resource wastage. In general, as seen by Figure~\ref{fig:T_graphs}, it may seem that by increasing $T$, we would be able to achieve a higher accuracy and resource efficiency. However, this is not ideal as it does not consider the fact that larger $T$ would imply longer round duration. In fact, a similar accuracy and resource efficiency can be achieved at $T=0.5$ when compared to $T=2$, but with $\frac{1}{4}$th the time spent on training. Therefore, $T=0.5$, might be a better solution in this particular scenario.

\subsection{Heterogeneous Client Schemes}

\noindent\textbf{Age Weighted Update (AWU).} Now we compare the performance of the AWU scheme with the MCU scheme. In this experiment, we consider a certain percentage of clients to be extremely biased clients (i.e., local datasets are biased towards a certain class or few data points). For a biased client, we sample a few data points from class 0 and replicate them in their local datasets to match the number of data points of unbiased clients. For unbiased clients, we uniformly assign a single class from classes 1 to 9 (client $k$ is assigned class $\{k \; \text{mod}\; 9\}+1$ ) and randomly sample a certain number of data points from the allocated class. In this experiment, we choose $Q(x) = (\min\{x,10\})^2$ as the age based weighting function. In here, we further assume that, these biased clients are faster and provide updates in every training round. The variation of accuracy with the percentage of biased clients, for MCU and AWU schemes, is shown in Table~\ref{tab:mcu_vs_awu}. As depicted in Table~\ref{tab:mcu_vs_awu}, when the number of biased clients is fairly  low, MCU with $M=1$ seems to perform comparatively better than the AWU scheme. But as the number of biased clients increases, the AWU scheme seems to maintain its accuracy, while MCU scheme exhibits a significant reduction in accuracy.

\begin{table}[]
    \centering
    \begin{tabular}{|c|c|c|}
    \hline
         Biased client \% & MCU Acc.~ ($M=1)$ & AWU Acc.  \\
         \hline
         5 & 0.812 & 0.744\\ 
         \hline
          10 & 0.788 & 0.762\\ 
         \hline
          15 & 0.488 & 0.734\\ 
         \hline
          20 & 0.46 & 0.747\\ 
         \hline
          30 & 0.1 & 0.668\\ 
         \hline
    \end{tabular}
    \caption{Accuracy of the MCU ($M=1$) and AWU schemes for $T=0.5$ with different percentages of biased clients after 1000 training rounds. \vspace{-4mm} }
    \label{tab:mcu_vs_awu}
\end{table}

\noindent\textbf{Aggregated Gradient Update (AGU).} Next, we evaluate the performance of the MCU scheme against the AGU scheme when $M>1$. In this experiment, the client datasets are created by choosing a random number of classes to be allocated for each client and from each allocated class a random number of data points are sampled. For the AGU scheme, we use a fixed learning rate as opposed to the decaying learning rate used in MCU. Empirical evidence shows that a decaying learning rate works better for MCU scheme while a fixed learning rate is better for AGU scheme. The variation of the model accuracy with $M$, for the two schemes, is shown in Table~\ref{tab:mcu_vs_agu}. As shown in Table~\ref{tab:mcu_vs_agu}, as $M$ increases the performance degradation of MCU becomes more apparent when compared to the AGU scheme. For smaller $M$ values, their performances look identical with MCU having a slight edge over AGU.

\begin{table}[]
    \centering
    \begin{tabular}{|c|c|c|}
    \hline
         $M$ & MCU Acc. ~ ($M>1)$ & AGU Acc.  \\
         \hline
         27 & 0.938 & 0.932\\ 
         \hline
          29 & 0.926 & 0.917\\ 
         \hline
          31 & 0.892 & 0.915\\ 
         \hline
          33 & 0.851 & 0.902\\ 
         \hline
          35 & 0.785 & 0.884\\ 
         \hline
    \end{tabular}
    \caption{Accuracy of the MCU  and AGU schemes for $T=0.3$ with $M$ after 1000 training rounds. \vspace{-4mm} }
    \label{tab:mcu_vs_agu}
\end{table}

\section{Conclusion}
In this work, we have uncovered how age is fundamentally associated with the model accuracy for the studied MCU scheme and have provided analytical methods to fine tune the system parameters to achieve the desired balance between resource efficiency and model performance. We have shown that, in general, updating with the available number of clients without waiting for a large number of clients can yield better model performance in terms of both model accuracy and convergence time. We have also introduced two schemes that can be used in the presence heterogeneous clients and have shown through numerical results, that these two schemes are more reliable in the presence of biased or adversarial clients.

\appendix
\section{Proof of Theorem \ref{thrm:mcu_conv}}\label{app:conv}
Let the SGD gradient $\tilde g_t$ and the actual gradient $ g_t$ relative to the entire client datasets, in the round $t$, be defined as follows,
\begin{align}
    \tilde g_t &= \frac{\ind_{\{|S_t|\geq M\}}}{|S_t|}\sum_{k\in S_t}\nabla\tilde F_k(w_t)\\
     g_t &= \frac{\ind_{\{|S_t|\geq M\}}}{|S_t|}\sum_{k\in S_t}\nabla F_k(w_t).
\end{align}
Then, as in \cite{fedavg_nidd}, the mean square error of the  global model in round $(t+1)$ is given by,
\begin{align}
    \e\|w_{t+1}-w^*\|^2&=\e\|w_t-\eta_t \tilde g_t-w^*+\eta_t g_t-\eta_tg_t\|^2 \nonumber\\
                        &=\e\|w_t-\eta_tg_t - w^*\|^2 + \eta_t^2\e\| g_t-\tilde g_t\|^2 \nonumber \\&\qquad+2\eta_t\e\langle w_t-\eta_t g_t-w^*, g_t-\tilde g_t\rangle. \label{eqn:mse_err}
\end{align}
Using Assumption~\ref{assum:ube} and tower property of expectation, we see that,
\begin{align}
    &\e\langle w_t-\eta_t g_t-w^*, g_t-\tilde g_t\rangle\nonumber\\
    &=\e[\langle w_t-\eta_t g_t-w^*,\e[ g_t-\tilde g_t|w_t,S_t]\rangle]\nonumber\\
    &=\e[\langle w_t-\eta_t g_t-w^*,\frac{\ind_{\{|S_t|\geq M\}}}{|S_t|}\sum_{k\in S_t}\e[\nabla F_k(w_t)-\nabla\tilde F_k(w_t)|w_t,S_t]\rangle] \nonumber\\
    &=0.
\end{align}
This reduces \eqref{eqn:mse_err}  to the following,
\begin{align}\label{eqn:mse_err_2}
     \e\|w_{t+1}-w^*\|^2&= \e\|w_t-\eta_t g_t - w^*\|^2 + \eta_t^2\e\| g_t- \tilde g_t\|^2.
\end{align}
Then, to find a convergence bound we only need to bound the terms $\e\|w_t-\eta_t g_t - w^*\|^2$ and $\e\| g_t-\tilde g_t\|^2$. The latter can be bounded as,
\begin{align}\label{eqn:gtilde_b}
    \e\| g_t-\tilde g_t\|^2
    &=\e\left[\|\frac{\ind_{\{|S_t|\geq M\}}}{|S_t|}\sum_{k\in S_t}(\nabla F_k(w_t)-\nabla\tilde F_k(w_t))\|^2\right]\nonumber\\
    &=\e\left[\frac{\ind_{\{|S_t|\geq M\}}}{|S_t|^2}\e\big[\|\sum_{k\in S_t}(\nabla F_k(w_t)-\nabla\tilde F_k(w_t))\|^2|S_t,w_t\big]\right] \nonumber\\
    &=\e\left[\frac{\ind_{\{|S_t|\geq M\}}}{|S_t|^2}\sum_{k=1}^N\ind_{\{k\in S_t\}}\e\|(\nabla F_k(w_t)-\nabla\tilde F_k(w_t))\|^2\right]\nonumber\\
    &\leq\e\left[\frac{\ind_{\{|S_t|\geq M\}}}{|S_t|^2}\sum_{k=1}^N\ind_{\{k\in S_t\}}\sigma^2\right]\nonumber\\
    &\leq \frac{\sigma^2}{M}\sum_{k=1}^N\e\left[\frac{\ind_{\{k\in S_t,|S_t|\geq M\}}}{|S_t|}\right]\nonumber\\
    &=\frac{\sigma^2 N\tilde S}{M}.
\end{align}
Here, $\tilde S = \e\left[\frac{\ind_{\{k\in S_t,|S_t|\geq M\}}}{|S_t|}\right]$, the third equality was obtained by Assumption~\ref{assum:ind} and first inequality using Assumption~\ref{assum:ube}.

The first term in \eqref{eqn:mse_err_2} is given by the following,
\begin{align}\label{eqn:mse_err_3}
    \e\|w_t-\eta_t g_t - w^*\|^2 &=\e\|w_t-w^*\|^2+\eta_t^2\e\| g_t\|^2\nonumber\\ & \qquad-2\eta_t\e\langle w_t-w^*, g_t\rangle
\end{align}
Using Jensen's inequality and Assumption~\ref{assum:smooth}, we can bound the second term in \eqref{eqn:mse_err_3} as follows,
\begin{align}\label{eqn:gtb}
    \e\|g_t\|^2 &= \e \|\frac{\ind_{\{|S_t|\geq M\}}}{|S_t|}\sum_{k\in S_t}\nabla F_k(w_t)\|^2 \nonumber\\
    &\leq \e\left[ \frac{\ind_{\{|S_t|\geq M\}}}{|S_t|}\sum_{k=1}^{N}\ind_{\{k \in S_t\}} \e \|\nabla F_k(w_t)\|^2 \right]\nonumber\\
    &\leq 2L\e\left[\sum_{k=1}^{N}\frac{\ind_{\{k\in S_t,|S_t|\geq M\}}}{|S_t|}(F_k(w_t)-F_k^*)\right]\nonumber\\
    &=2L \e\left[ \frac{\ind_{\{k\in S_t,|S_t|\geq M\}}}{|S_t|}\right]\e\left[\sum_{k=1}^{N}(F_k(w_t)-F_k^*)\right]\nonumber\\
    &=2L\tilde S \e\left[\sum_{k=1}^{N}(F_k(w_t)-F_k^*)\right].
\end{align}
The third term in \eqref{eqn:mse_err_3} can be bounded using Assumption~\ref{assum:conv} as follows,
\begin{align}\label{eqn:ipb}
    \e[\langle &w_t-w^*,g_t\rangle] \nonumber\\
    &= \e\left[ \sum_{k=1}^{N}\frac{\ind_{\{k \in S_t, |S_t|\geq M\}}}{|S_t|}\langle w_t-w^*,\nabla F_k(w_t)\rangle\right]\nonumber\\
    &=\e\left[\frac{\ind_{\{k \in S_t, |S_t|\geq M\}}}{|S_t|} \right]\sum_{k=1}^{N}\e\left[\langle w_t-w^*,\nabla F_k(w_t)\rangle\right]\nonumber\\
    &\geq \tilde S \sum_{k=1}^{N}\e\left[F_k(w_t)-F_k(w^*)+\frac{\mu}{2}\|w_t-w^*\|^2\right]. 
\end{align}
By using \eqref{eqn:mse_err_3}, \eqref{eqn:gtb} and \eqref{eqn:ipb}, we obtain,
\begin{align} \label{eqn:mse_err_3_inq}
    \e\|w_t&-\eta_t g_t - w^*\|^2\nonumber\\
    \leq &(1-\tilde S N \mu \eta_t) \e\|w_t-w^*\|^2+2L\tilde S \eta_t^2 \sum_{k=1}^N[\e[F_k(w_t)]-F_k^*]\nonumber\\
    &-2\eta_t\tilde S\sum_{k=1}^N[\e[F_k(w_t)]-F_k(w^*)].
\end{align}
Using the fact that $NF^* =\sum_{k=1}^{N}F_k(w^*)$  and selecting the learning rate such that $\eta_t\leq \frac{1}{L}$, we can bound the second and the third terms in the \eqref{eqn:mse_err_3_inq} as follows,
\begin{align}\label{eqn:niid_b}
    2L\tilde S & \eta_t^2 \sum_{k=1}^N[\e[F_k(w_t)]-F_k^*]-2\eta_t\tilde S\sum_{k=1}^N[\e[F_k(w_t)]-F_k(w^*)]\nonumber\\
    &= 2\tilde S \eta_t (L\eta_t -1) \sum_{k=1}^N[\e[F_k(w_t)]-F^*]+2L\tilde S \eta_t^2 \sum_{k=1}^N[F^*-F_k^*]\nonumber\\
    &\leq 2L\tilde S\Gamma N \eta_t^2,
\end{align}
where the inequality in \eqref{eqn:niid_b} is obtained using the fact that $[\e[F_k(w_t)]-F^*]\geq 0$ and $L\eta_t\leq1$ for $\eta_t\leq \frac{1}{L}$ while the term $\sum_{k=1}^N[F^*-F_k^*]=N\Gamma$ by definition.

Therefore, by combining \eqref{eqn:mse_err_2}, \eqref{eqn:gtilde_b}, \eqref{eqn:mse_err_3_inq} and \eqref{eqn:niid_b}, we obtain,
\begin{align} \label{eqn:opt_gap}
    \e\|&w_{t+1}-w^*\|^2\nonumber\\
    &\leq (1-\tilde S N\mu \eta_t)\e\|w_t-w^*\|^2+\eta_t^2(2\tilde S L N\Gamma+\sigma^2\tilde S \frac{N}{M})\nonumber\\
    &= (1-A\eta_t)\e\|w_t-w^*\|^2+\eta_t^2B,
\end{align}
where $A= \tilde S N \mu$ and $B= 2\tilde S L N \Gamma + \sigma^2\tilde S \frac{N}{M}$.
 
It is shown in \cite{fedavg_nidd} that, if the deviation of the model from the optimal point in round $(t+1)$ is upper bounded by the deviation of the model from the optimal point in round $t$ for some constants $A$ and $B$ as given in \eqref{eqn:opt_gap}, then by choosing $\eta_t = \frac{\beta}{\gamma + t}$ such that $\beta > \frac{1}{A}$, $\gamma>0$ and satisfies $\eta_1 \leq \min \{\frac{1}{A},\frac{1}{L}\}$, we can bound the optimality gap of round $t$ as follows,
\begin{align}
     \e[F(w_t)]-F^*\leq \frac{L}{2(\gamma+t)}\left[ \frac{\beta^2B}{\beta A -1}+(\gamma+1)\e\|w_1-w^*\|^2\right].
\end{align}
Therefore, by choosing $\beta = \frac{2}{A}$ and $\gamma =\frac{2L}{A}+1$, we get the following,
\begin{align}
     \e[F(w_t)]-F^*&\leq \frac{AL}{2(2L+A(t+1))}\left[\frac{4B}{A^2}+\frac{2(L+A)}{A}\varepsilon_1\right]\nonumber\\
     &=\frac{L}{(2L+A(t+1))}\left[\frac{2B}{A}+(L+A)\varepsilon_1\right]\nonumber\\
     &=\frac{L}{(2L+\tilde S N\mu t)}\left[\frac{(4L\Gamma+\frac{2\sigma^2}{M})}{\mu}+(L+N\mu\tilde S)\varepsilon_1\right], \label{final-eq}
\end{align}
where, in the last equality, we have approximated $t+1$ with $t$. The final expression in \eqref{final-eq} is the same as \eqref{eqn:mcu_conv}, concluding the proof.

\bibliographystyle{ACM-Reference-Format} 
\bibliography{refs}


\begin{thebibliography}{25}

%%% the Web address.

\ifx \showCODEN    \undefined \def \showCODEN     #1{\unskip}     \fi
\ifx \showDOI      \undefined \def \showDOI       #1{#1}\fi
\ifx \showISBNx    \undefined \def \showISBNx     #1{\unskip}     \fi
\ifx \showISBNxiii \undefined \def \showISBNxiii  #1{\unskip}     \fi
\ifx \showISSN     \undefined \def \showISSN      #1{\unskip}     \fi
\ifx \showLCCN     \undefined \def \showLCCN      #1{\unskip}     \fi
\ifx \shownote     \undefined \def \shownote      #1{#1}          \fi
\ifx \showarticletitle \undefined \def \showarticletitle #1{#1}   \fi
\ifx \showURL      \undefined \def \showURL       {\relax}        \fi
% invisible to TeX
\providecommand\bibfield[2]{#2}
\providecommand\bibinfo[2]{#2}
\providecommand\natexlab[1]{#1}
\providecommand\showeprint[2][]{arXiv:#2}

\bibitem[A.~Kosta and Angelakis(2017)]%
        {age1}
\bibfield{author}{\bibinfo{person}{N.~Pappas A.~Kosta} {and} \bibinfo{person}{V. Angelakis}.} \bibinfo{year}{2017}\natexlab{}.
\newblock \showarticletitle{Age of information: A new concept, metric, and tool.}
\newblock \bibinfo{journal}{\emph{Foundations and Trends in Networking}}  \bibinfo{volume}{12} (\bibinfo{date}{November} \bibinfo{year}{2017}), \bibinfo{pages}{162--259}.
\newblock


\bibitem[Abdelmoniem et~al\mbox{.}(2023)]%
        {refl}
\bibfield{author}{\bibinfo{person}{A.~M. Abdelmoniem}, \bibinfo{person}{A.~N.~Sahu andM. Canini}, {and} \bibinfo{person}{S.~A. Fahmy}.} \bibinfo{year}{2023}\natexlab{}.
\newblock \showarticletitle{REFL: Resource-Efficient Federated Learning}. In \bibinfo{booktitle}{\emph{EUROSYS}}.
\newblock


\bibitem[Buyukates and Ulukus(2021)]%
        {buyukates_2021}
\bibfield{author}{\bibinfo{person}{B. Buyukates} {and} \bibinfo{person}{S. Ulukus}.} \bibinfo{year}{2021}\natexlab{}.
\newblock \showarticletitle{Timely Communication in Federated Learning}. In \bibinfo{booktitle}{\emph{IEEE Infocom}}.
\newblock


\bibitem[Chen et~al\mbox{.}(2021)]%
        {dl1}
\bibfield{author}{\bibinfo{person}{C. Chen}, \bibinfo{person}{H. Xu}, \bibinfo{person}{W. Wang}, \bibinfo{person}{B. Li}, \bibinfo{person}{Bo Li}, \bibinfo{person}{L. Chen}, {and} \bibinfo{person}{G. Zhang}.} \bibinfo{year}{2021}\natexlab{}.
\newblock \showarticletitle{Communication-Efficient Federated Learning with Adaptive Parameter Freezing}. In \bibinfo{booktitle}{\emph{IEEE ICDCS}}.
\newblock


\bibitem[Damaskinos et~al\mbox{.}(2022)]%
        {fleet}
\bibfield{author}{\bibinfo{person}{G. Damaskinos}, \bibinfo{person}{R. Guerraoui}, \bibinfo{person}{A.-M. Kermarrec}, \bibinfo{person}{V. Nitu}, \bibinfo{person}{R. Patra}, {and} \bibinfo{person}{F. Taiani}.} \bibinfo{year}{2022}\natexlab{}.
\newblock \showarticletitle{FLeet: Online Federated Learning via Staleness Awareness and Performance Prediction}.
\newblock \bibinfo{journal}{\emph{ACM Trans. on Intell. Sys. and Tech.}} \bibinfo{volume}{13}, \bibinfo{number}{5} (\bibinfo{date}{September} \bibinfo{year}{2022}), \bibinfo{pages}{1--30}.
\newblock


\bibitem[Den(2012)]%
        {mnist}
\bibfield{author}{\bibinfo{person}{L. Den}.} \bibinfo{year}{2012}\natexlab{}.
\newblock \showarticletitle{The MNIST Database of Handwritten Digit Images for Machine Learning Research [Best of the Web]}.
\newblock \bibinfo{journal}{\emph{IEEE Signal Processing Magazine,}} \bibinfo{volume}{29}, \bibinfo{number}{6} (\bibinfo{date}{October} \bibinfo{year}{2012}), \bibinfo{pages}{141--142}.
\newblock


\bibitem[Feng et~al\mbox{.}(2020)]%
        {pmf}
\bibfield{author}{\bibinfo{person}{J. Feng}, \bibinfo{person}{C. Rong}, \bibinfo{person}{F. Sun}, \bibinfo{person}{D. Guo}, {and} \bibinfo{person}{Y. Li}.} \bibinfo{year}{2020}\natexlab{}.
\newblock \showarticletitle{PMF: a privacy-preserving human mobility prediction framework via federated learning}.
\newblock \bibinfo{journal}{\emph{ACM Interactive, Mobile, Wearable and Ubiquitous Technologies}} \bibinfo{volume}{4}, \bibinfo{number}{1} (\bibinfo{date}{March} \bibinfo{year}{2020}), \bibinfo{pages}{10--21}.
\newblock


\bibitem[Kaul et~al\mbox{.}(2012)]%
        {rts2012}
\bibfield{author}{\bibinfo{person}{S.~K. Kaul}, \bibinfo{person}{R.~D. Yates}, {and} \bibinfo{person}{M. Gruteser}.} \bibinfo{year}{2012}\natexlab{}.
\newblock \showarticletitle{Real-time status: How often should one update?}. In \bibinfo{booktitle}{\emph{IEEE Infocom}}.
\newblock


\bibitem[Konencny et~al\mbox{.}(2016)]%
        {fedavg}
\bibfield{author}{\bibinfo{person}{J. Konencny}, \bibinfo{person}{H.~B. McMahan}, \bibinfo{person}{F.~X. Yu}, \bibinfo{person}{P. Richtarik}, \bibinfo{person}{A.~T. Suresh}, {and} \bibinfo{person}{D. Bacon}.} \bibinfo{year}{2016}\natexlab{}.
\newblock \showarticletitle{Federated learning: Strategies for improving communication efficiency}.
\newblock \bibinfo{journal}{\emph{ACM Interact. Mob. Wearable Ubiq. Tech}} (\bibinfo{year}{2016}).
\newblock
\newblock
\shownote{Available online at arXiv:1610.05492}.


\bibitem[Lai et~al\mbox{.}(2021)]%
        {oort}
\bibfield{author}{\bibinfo{person}{F. Lai}, \bibinfo{person}{X. Zhu}, \bibinfo{person}{H.~V. Madhyastha}, {and} \bibinfo{person}{M. Chowdhury}.} \bibinfo{year}{2021}\natexlab{}.
\newblock \showarticletitle{Efficient Federated Learning via Guided Participant Selection}. In \bibinfo{booktitle}{\emph{USENIX Symposium on Operating Systems Design and Implementation}}.
\newblock


\bibitem[Li and Gong(2023)]%
        {anarch}
\bibfield{author}{\bibinfo{person}{D. Li} {and} \bibinfo{person}{X. Gong}.} \bibinfo{year}{2023}\natexlab{}.
\newblock \showarticletitle{Anarchic Federated learning with Delayed Gradient Averaging}. In \bibinfo{booktitle}{\emph{MOBIHOC}}.
\newblock


\bibitem[Li et~al\mbox{.}(2020a)]%
        {fed_apps}
\bibfield{author}{\bibinfo{person}{L. Li}, \bibinfo{person}{Y. Fan}, \bibinfo{person}{M. Tse}, {and} \bibinfo{person}{K.-Y. Lin}.} \bibinfo{year}{2020}\natexlab{a}.
\newblock \showarticletitle{A review of applications in federated learning}.
\newblock \bibinfo{journal}{\emph{Elsevier Computers \& Industrial Engineering}}  \bibinfo{volume}{149} (\bibinfo{date}{November} \bibinfo{year}{2020}), \bibinfo{pages}{1--15}.
\newblock


\bibitem[Li et~al\mbox{.}(2020b)]%
        {fedavg_nidd}
\bibfield{author}{\bibinfo{person}{X. Li}, \bibinfo{person}{K. Huang}, \bibinfo{person}{W. Yang}, \bibinfo{person}{S. Wang}, {and} \bibinfo{person}{Z. Zhang}.} \bibinfo{year}{2020}\natexlab{b}.
\newblock \showarticletitle{On the Convergence of FedAvg on Non-IID Data}. In \bibinfo{booktitle}{\emph{ICLR}}.
\newblock


\bibitem[Liu et~al\mbox{.}(2021)]%
        {age-aware}
\bibfield{author}{\bibinfo{person}{X. Liu}, \bibinfo{person}{X. Qin}, \bibinfo{person}{H. Chen}, \bibinfo{person}{Y. Liu}, \bibinfo{person}{B. Liu}, {and} \bibinfo{person}{P. Zhang}.} \bibinfo{year}{2021}\natexlab{}.
\newblock \showarticletitle{Age-aware Communication Strategy in Federated Learning with Energy Harvesting Devices}. In \bibinfo{booktitle}{\emph{IEEE ICCC}}.
\newblock


\bibitem[McMahan et~al\mbox{.}(2017)]%
        {McMahan}
\bibfield{author}{\bibinfo{person}{H.~B. McMahan}, \bibinfo{person}{E. Moore}, \bibinfo{person}{D. Ramage}, \bibinfo{person}{S. Hampson}, {and} \bibinfo{person}{B.~Aguera y Areas}.} \bibinfo{year}{2017}\natexlab{}.
\newblock \showarticletitle{Communication-efficient learning of deep networks from decentralized data}. In \bibinfo{booktitle}{\emph{AISTATS}}.
\newblock


\bibitem[Mitra and Ulukus(2023)]%
        {mitra_2023}
\bibfield{author}{\bibinfo{person}{P. Mitra} {and} \bibinfo{person}{S. Ulukus}.} \bibinfo{year}{2023}\natexlab{}.
\newblock \showarticletitle{Timely Asynchronous Hieracrchical Federated Learning: Age of Convergence}. In \bibinfo{booktitle}{\emph{WiOpt}}.
\newblock


\bibitem[Sun et~al\mbox{.}(2019)]%
        {age2}
\bibfield{author}{\bibinfo{person}{Y. Sun}, \bibinfo{person}{I.Kadota}, \bibinfo{person}{R. Talak}, {and} \bibinfo{person}{E.~H. Modiano}.} \bibinfo{year}{2019}\natexlab{}.
\newblock \showarticletitle{Age of information: A new metric for information freshness.}
\newblock \bibinfo{journal}{\emph{Age of information}}  \bibinfo{volume}{12} (\bibinfo{date}{December} \bibinfo{year}{2019}), \bibinfo{pages}{1--224}.
\newblock


\bibitem[Wu et~al\mbox{.}(2020)]%
        {safa}
\bibfield{author}{\bibinfo{person}{W. Wu}, \bibinfo{person}{L. He}, \bibinfo{person}{W. Lin}, \bibinfo{person}{R. Mao}, \bibinfo{person}{C. Maple}, {and} \bibinfo{person}{S. Jarvis}.} \bibinfo{year}{2020}\natexlab{}.
\newblock \showarticletitle{SAFA: A Semi-Asynchronous Protocol for Fast Federated Learning with Low Overhead}.
\newblock \bibinfo{journal}{\emph{IEEE Trans. Comput,}} \bibinfo{volume}{70}, \bibinfo{number}{5} (\bibinfo{date}{May} \bibinfo{year}{2020}), \bibinfo{pages}{655--668}.
\newblock


\bibitem[Xie et~al\mbox{.}(2019)]%
        {async1}
\bibfield{author}{\bibinfo{person}{C. Xie}, \bibinfo{person}{S. Koyejo}, {and} \bibinfo{person}{I. Gupta}.} \bibinfo{year}{2019}\natexlab{}.
\newblock \showarticletitle{Asynchronous federated optimization}.
\newblock  (\bibinfo{year}{2019}).
\newblock
\newblock
\shownote{Available online at arXiv:1903.03934}.


\bibitem[Xu et~al\mbox{.}(2021)]%
        {async2}
\bibfield{author}{\bibinfo{person}{C. Xu}, \bibinfo{person}{Y. Qu}, \bibinfo{person}{Y. Xiang}, {and} \bibinfo{person}{L. Gao}.} \bibinfo{year}{2021}\natexlab{}.
\newblock \showarticletitle{Asynchronous federated learning on heterogeneous devices: A survey}.
\newblock  (\bibinfo{year}{2021}).
\newblock
\newblock
\shownote{Available online at arXiv:2109:04269}.


\bibitem[Xue et~al\mbox{.}(2022)]%
        {cost_aware}
\bibfield{author}{\bibinfo{person}{D. Xue}, \bibinfo{person}{J. Luo}, \bibinfo{person}{C. Jiang}, {and} \bibinfo{person}{L. Gao}.} \bibinfo{year}{2022}\natexlab{}.
\newblock \showarticletitle{Cost-Awaare Hierarchical Federated Learning via Over-the-Air Computing}. In \bibinfo{booktitle}{\emph{IEEE Globecom}}.
\newblock


\bibitem[Yang et~al\mbox{.}(2020)]%
        {arafa2020}
\bibfield{author}{\bibinfo{person}{H.~H. Yang}, \bibinfo{person}{A. Arafa}, \bibinfo{person}{T.~Q.~S. Quek}, {and} \bibinfo{person}{H.~V. Poor}.} \bibinfo{year}{2020}\natexlab{}.
\newblock \showarticletitle{Age-Based Scheduling Policy for Federated Learning in Mobile Edge Networks}. In \bibinfo{booktitle}{\emph{IEEE ICASSP}}.
\newblock


\bibitem[Yates et~al\mbox{.}(2020)]%
        {age3}
\bibfield{author}{\bibinfo{person}{R.~D. Yates}, \bibinfo{person}{Y. Sun}, \bibinfo{person}{D.~R. Brown}, \bibinfo{person}{S.~K. Kaul}, \bibinfo{person}{E. Modiano}, {and} \bibinfo{person}{S. Ulukus}.} \bibinfo{year}{2020}\natexlab{}.
\newblock \showarticletitle{Age of information: An introduction and survey.}
\newblock \bibinfo{journal}{\emph{IEEE Jour. Sel. Areas in Comm,}} \bibinfo{volume}{39}, \bibinfo{number}{5} (\bibinfo{date}{May} \bibinfo{year}{2020}), \bibinfo{pages}{1183--1210}.
\newblock


\bibitem[Zhang et~al\mbox{.}(2023)]%
        {FedTune}
\bibfield{author}{\bibinfo{person}{H. Zhang}, \bibinfo{person}{L. Fu}, \bibinfo{person}{M. Zhang}, \bibinfo{person}{P. Hu}, \bibinfo{person}{X. Cheng}, \bibinfo{person}{P. Mohapatra}, {and} \bibinfo{person}{X. Liu}.} \bibinfo{year}{2023}\natexlab{}.
\newblock \showarticletitle{Federated Learning Hyperparameter Tuning From a System Perspective}.
\newblock \bibinfo{journal}{\emph{IEEE Internet of Things Journal,}} \bibinfo{volume}{10}, \bibinfo{number}{16} (\bibinfo{date}{March} \bibinfo{year}{2023}), \bibinfo{pages}{14102--14113}.
\newblock


\bibitem[Zhang et~al\mbox{.}(2022)]%
        {dl2}
\bibfield{author}{\bibinfo{person}{Y. Zhang}, \bibinfo{person}{L. Duan}, {and} \bibinfo{person}{N.-M Cheung}.} \bibinfo{year}{2022}\natexlab{}.
\newblock \showarticletitle{Accelerating Federated Learning on Non-IID Data Against Stragglers}. In \bibinfo{booktitle}{\emph{IEEE SECON}}.
\newblock


\end{thebibliography}

\end{document}